\documentclass[pdflatex,sn-mathphys-num]{sn-jnl}

\usepackage{graphicx}%
\usepackage{multirow}%
\usepackage{amsmath,amssymb,amsfonts}%
\usepackage{amsthm}%
\usepackage{mathrsfs}%
\usepackage[title]{appendix}%
\usepackage{xcolor}%
\usepackage{textcomp}%
\usepackage{manyfoot}%
\usepackage{booktabs}%
\usepackage{algorithm}%
\usepackage{algorithmicx}%
\usepackage{algpseudocode}%
\usepackage{listings}%
\usepackage{subcaption}
\usepackage{hyperref}

\theoremstyle{thmstyleone}%
\theoremstyle{thmstyletwo}%

\theoremstyle{thmstylethree}%

\title[Article Title]{How to Efficiently Annotate Images for Best-Performing Deep Learning-Based Segmentation Models: An Empirical Study with Weak and Noisy Annotations and Segment Anything Model}

\begin{document}

\author*[1]{\fnm{Yixin}\sur{Zhang}}\email{yixin.zhang7@duke.edu}
\author[1]{\fnm{Shen}\sur{Zhao}}
\author[1]{\fnm{Hanxue}\sur{Gu}}\email{hanxue.gu@duke.edu}
\author[1,2,3,4]{\fnm{Maciej A.}\sur{Mazurowski}}\email{maciej.mazurowski@duke.edu}

\affil[1]{\orgdiv{Department of Electrical and Computer Engineering}, \orgname{Duke University}, \orgaddress{\city{Durham}, \country{USA}}}

\affil[2]{\orgdiv{Department of Radiology}, \orgname{Duke University}, \orgaddress{\city{Durham}, \country{USA}}}

\affil[3]{\orgdiv{Department of Biostatistics \& Bioinformatics}, \orgname{Duke University}, \orgaddress{\city{Durham}, \country{USA}}}

\affil[4]{\orgdiv{Department of Computer Science}, \orgname{Duke University}, \orgaddress{\city{Durham}, \country{USA}}}



\abstract{Deep neural networks (DNNs) have demonstrated exceptional performance across various image segmentation tasks. However, the process of preparing datasets for training segmentation DNNs is both labor-intensive and costly, as it typically requires pixel-level annotations for each object of interest. To mitigate this challenge, alternative approaches such as using weak labels (e.g., bounding boxes or scribbles) or less precise (noisy) annotations can be employed. Noisy and weak labels are significantly quicker to generate, allowing for more annotated images within the same time frame. However, the potential decrease in annotation quality may adversely impact the segmentation performance of the resulting model. In this study, we conducted a comprehensive cost-effectiveness evaluation on {six} variants  of annotation strategies ({9$\sim$10} sub-variants in total) across 4 datasets, {and} conclude that the common practice of precisely outlining objects of interest is {virtually} never the optimal approach when annotation {budget} is limited. {Both noisy and weak annotations showed usage cases that yield similar performance to the perfectly annotated counterpart, yet had significantly better cost-effectiveness.}   We hope our findings will help researchers {be aware of the different available options and}  use their annotation budgets more efficiently, especially in cases where accurately acquiring labels for target objects is particularly costly. Our code will be made available on \url{https://github.com/yzluka/AnnotationEfficiency2D}}




\maketitle
\section*{Disclaimer}

\textcolor{red}{
This paper is a preprint and has not been peer-reviewed. The published final form may be accessed at \url{https://doi.org/10.1007/s10278-025-01408-7} }

\section{Background}\label{bg}
Semantic segmentation is a well-researched topic in the field of computer vision. 
It has found a variety of practical applications.
Segmentation neural networks typically employ Convolutional Neural Networks (CNNs) and/or Visual Transformers (ViTs) architectures. 
A notable challenge in training such models is the availability of segmentation labels, which typically comprise pixel-level segmentation masks. 
This is particularly the case where annotating data requires considerable professional expertise such as medical imaging.

Two main branches of solutions emerged as remedies to alleviate the issue of limited precise segmentation labels. One is semi-supervised learning (SSL), which uses both annotated and unannotated data in the training process.
Some common approaches include training feature extractors with proxy tasks such as self-interpolation \cite{MAE,SimMiM}, consistency regularization for augmentation \cite{ouali2020semi,chen2021semi}, and/or contrastive learning \cite{chen2020simple}.
Fine-tuning then follows on the task-relevant fully annotated images. The other is weakly-supervised semantic segmentation (WSSS), where imperfect labels are used to train the model \cite{papandreou2015weakly}. 
WSSS trades annotation quality for more annotated images.
Given annotations assumed in WSSS were not actual attempts to follow the boundaries of the objects of interest (in contrast to annotation with error/noise), they are also called “weak annotations”. Examples of weak annotations include points, scribbles, and bounding boxes. 

Directly training models with weak annotations such as bounding boxes, points or scribbles rarely results in desirable performance. To employ them to train a model, some label pre-processing or modifications of training frameworks are needed.
For bounding box annotations, studies \emph{dated before 2018} such as Multiscale Combinatorial Grouping (MCG) \cite{arbelaez2014multiscale}, Simply Does It (SDI) \cite{khoreva2017simple}, and GrabCut \cite{rother2004grabcut} generate pseudo-masks for objects with probabilistic graphical algorithms, and then use the generated masks to train the neural network.
Subsequent work such as Box-drive Class-wise region Masking (BCM) \cite{song2019box} and Background Aware Pooling (BAP) \cite{oh2021background} favor applying training-time regularization and post-processing based on Conditional Random Field (CRF) \cite{krahenbuhl_efficient_2011} for iterative mask updates, rather than pre-generating the masks before the model training. 
For scribbles and point annotations, Can et al. \cite{can2018learning} iteratively update the pseudo label based on some form of prediction uncertainty measures, while Tange et al. \cite{tang2018regularized} train models with specifically designed loss functions for sparse annotations.
{Recently, aside from designing regularization terms to leverage properties of the specific category of objects \cite{GUO2023102790}}, models designed for interactive clicking segmentation such as RITM \cite{sofiiuk2022reviving}, Focal Click \cite{chen2022focalclick}
, or zero-short segmentation such as Segment Anything \cite{kirillov2023segment} start to gain popularity.

A related but slightly different concept from weak annotation is noisy annotation. 
{Unlike} weak annotations, {which} provide an abstract reference to the object, noisy annotations {outline the majority of the actual pixels associated with the target objects and exclude those irrelevant ones.}
Existing literature highlights some interesting 
characteristics of imprecision in segmentation annotations. Heller et al. \cite{heller2018imperfect} demonstrated that segmentation networks experience uniform performance degradation as label errors increase, with non-boundary-localized errors being less detrimental than boundary-localized ones. 
Vorontsov and Kadoury \cite{vorontsov2021label} showed segmentation networks to be resistant to varying degrees of random, unbiased errors.
Zlateski et al. \cite{zlateski2018importance} showed that models trained on datasets with a higher number of imperfect labels are more cost-effective than those with fewer precise labels when the training budget is limited. 
However, their research did not {include} weak annotations as potential annotation styles. 

In our study, we {experiment on} both weak annotations and noisy annotations under the {umbrella} of imperfect annotations. {Although many studies \cite{ren2020ufo,reiss2023decoupled,gao2022segmentation,wang2022omni} have analyzed methods and algorithms for integrating weak and noisy annotations into various downstream tasks, there has been limited comparative studies whose focuses are on the choices of annotation styles. Our study specifically addresses this gap for semantic segmentation tasks by empirically evaluating the cost-effectiveness of several commonly used weak and noisy annotation methods:}  1) polygons, 2) coarse contours, 3) bounding boxes, 4) scribbles 5) points, and 6) precise annotations in our study. {Some past publications \cite{7450177,8482248} also explore some less common weak and noisy annotation styles. These annotation styles are of very different nature from the ones include in our list and are beyond the scope of our study.}
Our analysis offers a guide for those developing segmentation models on how to allocate their limited time for creating annotations. 
It also helps determine whether and to what extent existing weak annotations can be effectively utilized as labels for semantic segmentation.

\section{Methods}\label{sec2}
Our study {consists of} three main steps: Annotation Time and Quality Estimation, Programmatic Imperfect Annotation Generation, and Annotation-Budget-Restricted Training. These steps are sequentially dependent, with the first two determining the quantity and characteristics of annotations used in the final step. In the last step, we control the annotation budget and compare different annotation types to {interprete} the most cost-effective approach for {each} segmentation task.

\subsection{Annotation Time and Quality Estimation}
{In this study, we use quotes requested from ten different image annotation service providers to estimate the cost ratio of among different annotation styles. The quality of noisy annotation quality are estimated from data collected from human participants in our preliminary study. We assumed a consistent cost ratio and mask quality across all datasets for the different annotation styles. We consider four datasets: two in the natural domain, focusing on "object" (VOC2012) \cite{pascal-voc-2012,BSDS500} and "stuff" (WATER) \cite{zhou2017scene,zhou2019semantic,satelliteWater}, respectively, and two in the medical domain, comprising CT data (LiTS17) \cite{bilic2023liver} and MRI data (BraTS20).} 
\begin{table}[h]
\centering
\begin{tabular}{||c|c|c|c|c|c|c||} 
 \hline
 &{precise} &	{polygon} &{coarse contours} &	{bounding box} &{ scribble} & {point}\\ [0.5ex] 
 \hline\hline
{Relative Cost} & {1.000} & {0.3018} & {0.3018} & {0.0846} & {0.0614} & {0.0559} \\ 
 \hline
{Mask Quality (IoU)}& {1.00}&{0.87} & {0.87}& {N/A}& {N/A}& {N/A}\\
 \hline
\end{tabular}
\caption{{Cost Ratio and Quality of Different Annotation Styles}}
\label{mask_cost_and_quality}
\end{table}

Since our experiments involve training models using various types of imperfect annotations, it is essential to generate realistic imperfect annotations algorithmically from precise ones. These generated annotations must visually and qualitatively approximate those created by human annotators. To accomplish this, we employed algorithms with adjustable parameters and optimized these parameters to ensure that the intersect-over-union score (IoU) between the precise annotation and the computer-generated imperfect annotation is comparable to that between the precise annotation and human-created imperfect annotation. The IoU score for the various types of annotations between human-created imperfect annotations and the precise annotation are listed in {Table \ref{mask_cost_and_quality}.}

\subsection{Programmatic Imperfect Annotation Generation}
\begin{figure}[ht!]
    \centering
    \begin{subfigure}[b]{0.32\textwidth}
        \centering
        \includegraphics[width=\textwidth]{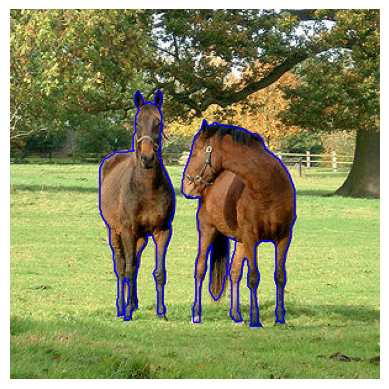} 
        \caption{Precise (prec)}
        \label{voc_pref}
    \end{subfigure}
    \begin{subfigure}[b]{0.32\textwidth}
        \centering
        \includegraphics[width=\textwidth]{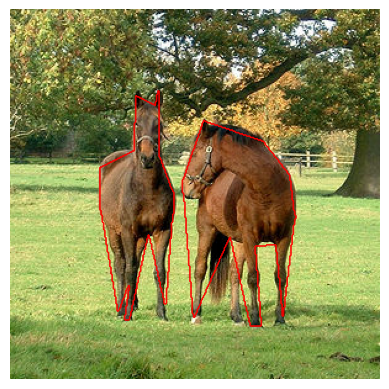} 
        \caption{Polygon (poly)}
        \label{voc_poly}
    \end{subfigure}
    \begin{subfigure}[b]{0.32\textwidth}
        \centering
        \includegraphics[width=\textwidth]{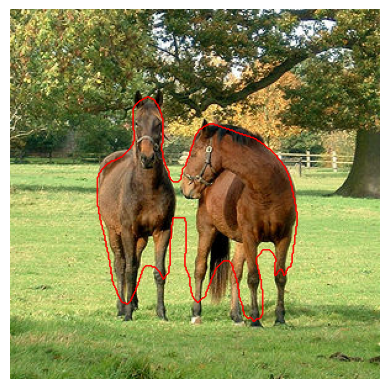} 
        \caption{Coarse Contour (rough)}
        \label{voc_rough}
    \end{subfigure}
    \vspace{-1mm}
    \begin{subfigure}[b]{0.32\textwidth}
        \centering
        \includegraphics[width=\textwidth]{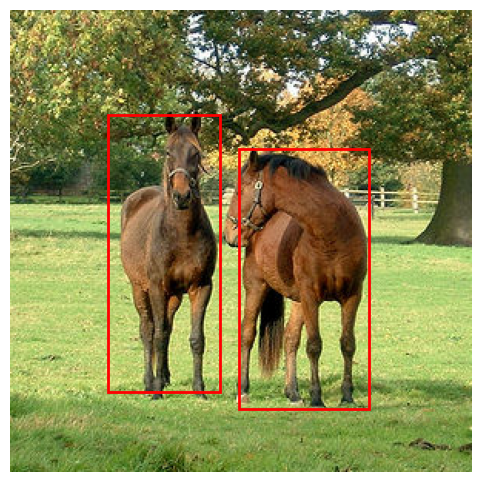} 
        \caption{Bounding Box (bbox)}
        \label{voc_bbox}
    \end{subfigure}
    \begin{subfigure}[b]{0.32\textwidth}
        \centering
        \includegraphics[width=\textwidth]{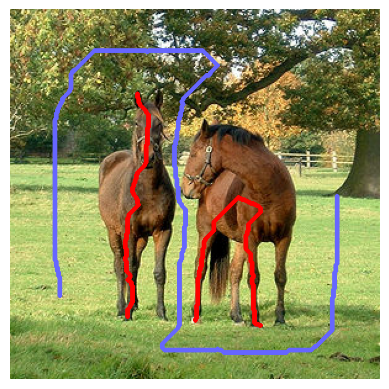} 
        \caption{Scribble (scrib)}
        \label{voc_scrib}
    \end{subfigure}
    \begin{subfigure}[b]{0.32\textwidth}
        \centering
        \includegraphics[width=\textwidth]{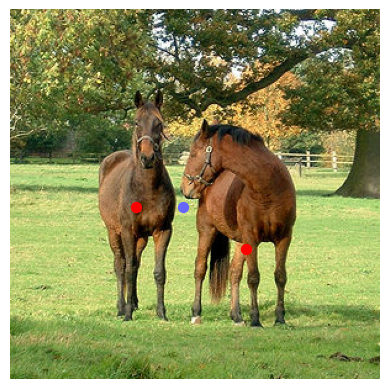} 
        \caption{Point (point)}
        \label{voc_point}
    \end{subfigure}
    \caption{Computer-generated annotations that assimilate {the six }different
annotation styles {on VOC2012. Blue contour indicates precise annotation, red contour/scribbles/point/bounding boxes indicate target objects, while the other color indicate background.}}
    \label{VOC_annot_samples}
\end{figure}

In this subsection, we reiterate the seven annotation styles involved in our study and provide a detailed account of our protocols to algorithmically generating imperfect annotations from the precise annotations. The seven annotation styles may be automatically generated following the recipe below:
\begin{enumerate}
    \item Precise(prec): These are the original labels provided with the dataset.
    \item Polygon (poly): Polygon labels are generated by approximating the boundaries of precise labels using the Ramer-Douglas-Peucker (RDP) algorithm. This process produces polygonal chains with vertices aligned along the boundaries of the precise annotation, enclosing a region with significant overlap.
    \item Coarse contour (rough)\footnote{Quotes and demos for coarse contour annotation are not provided by any annotation service provider, but we think it might be a good addition to consider. We assume coarse contour annotation consumes have the same average cost and similar quality to polygon masks.}: Coarse contour annotations simulate the outcome of an annotator quickly tracing along the boundaries of the objects. {These annotations are generated by first applying Gaussian blur to the segmentation mask and then retaining the pixels with intensities above a specified threshold.} Unlike {polygon}, which typically preserve the Hausdorff distance relative to the precise annotations, {coarse contour annotations} focus primarily on preserving the overall mass of the object. 
    Consequently, thinner and sharper features, such as spikes and small holes, may be omitted.
    \item Bounding boxe (bbox): Bounding box annotations are obtained by identifying the minimum and maximum horizontal and vertical coordinates of all ROI pixels in the precise annotation. {In practice, obtaining tight bounding boxes is challenging, so we followed Kirillov et al.'s approach \cite{kirillov2023segment} to introduce noise to the bounding boxes. This is done by adding random noise with a standard deviation equal to 10\% of the box's side length, capped at a maximum of 20 pixels, to each coordinate of the tight bounding box.}
    \item Scribbles (scrib): The process of generating scribble annotations adheres to the auto-scribble generation protocols outlined by Valvano et al. \cite{valvano2021learning}. Scribbles are generated by eroding the pre-existing precise segmentation masks using a standard skeletonization procedure. This process involves iteratively applying erosion until the target object loses connectivity in the next iteration. The longest path along the resulting "skeleton" is then identified as the scribble annotation, providing a visually realistic simulation of human-generated scribbles. One scribble is created for each nonconsecutive region and the background. {When using deriving points from scribbles as inputs to SAM, to obtain pseudo masks with highest-possible quality, we randomly sample between 7 and 15 points from each scribble segment to avoid under-segmentation for VOC2012 and WATER, as these two datasets are prone to under-segmentation; for the LiTS17 and BraTS20 dataset, to mitigating over-segmentation, we divide the scribble pixels into 1 $\sim$ 7 clusters with K-Mean Clustering based on number of scribble length relative to image size. Coordinates for pixels closest to the centroid are then derived as point-prompt.}
    \item Points (point): Point annotations can be considered a special case of scribbles, where each scribble is reduced to a single pixel. In implementation, we select the pixel closest to the centroid of the scribble for each contiguous segment of interest. The background region is also assigned a single point, even if it contains multiple disconnected pieces. A potential drawback of this approach is that a group of N people may be represented by only one point instead of N points.
\end{enumerate}
Fig. \ref{VOC_annot_samples} provided a {visualization of the} precise annotation along with the {five} computer-generated imperfect annotation styles involved in our study for semantic segmentation.
\subsection{Annotation-Budget-Restricted Training}
To identify the most cost-effective annotation type for segmentation tasks, we assess model performance across various time budgets on different annotation styles. Specifically, the budget levels are incremented by a factor of two until all available images in the training set can be precisely annotated. If the budget exceeds the number of available images, we train the model with all the labeled data available. At the conclusion of the experiments, we aim to produce a 2D plot with the annotation budget on the x-axis and the corresponding model performance (mean Intersection over Union (mIoU) or IoU of the target class) on the y-axis.

For each dataset, we utilize different model architectures and training configurations to explore a range of hyperparameter settings. {The model architectures and hyperparameters (e.g., batch size, learning rate, learning rate scheduling policies, and loss weighting) are selected to ensure that the results we obtain are comparable to those reported in the benchmark records, assuming the full precisely annotated datasets were used.} This training recipe remains unchanged across all combinations of annotation budget and annotation style. Early stopping is used when models are trained with a subset of the dataset or imperfect annotations, as models generally converge faster under these conditions.

\subsubsection{Model Architecture and Data Augmentation}
We paired four different model architectures with the four datasets.
\begin{enumerate}
    \item For VOC2012\footnote{contains a total of 12031 images with segmentation masks for a total of 20 classes of common objects. The ground-truth annotations for 2913 images are from VOC2012  while the rest are from BSDS500 \cite{BSDS500}}., we follow the official PSP-Net \cite{zhao2017pyramid} implementation with a ResNet-101 backbone and adhere to their training protocol. Data augmentation includes random mirroring, random resized cropping (between 0.5 and 2 times the original size), random rotations (from -10 to 10 degrees), and random Gaussian blur. This training setup is detailed on GitHub. Due to GPU memory constraints, we reduced the input patch size from 473 to 393 pixels. 
\item For WATER\footnote{WATER contains a total of 4070 images of water bodies, among which 1888 samples are from the water class in ADE20k \cite{zhou2017scene,zhou2019semantic}, 404 samples are from Water Segmentation Dataset \cite{liang2020waternet}, and 1778 samples are from Satellite Images of Water Bodies \cite{satelliteWater}}, we employ DeepLabV3 \cite{chen2017deeplab} with a ResNet-50 backbone pre-trained on ImageNet. Images are augmented with ColorJitter and RandomEqualize with a probability of 0.5, then padded to square and resized to $256 \times 256$ pixels. 
\item For LiTS17\footnote{LiTS17 contains 5501 2D slices from CT scans of 130 patients' livers. We simplify this dataset to have only two classes: the foreground (liver + tumor on the liver) and the background (non-liver pixels) so that the foreground aligns better with the concept of “object”.}, we use a vanilla U-Net \cite{ronneberger2015u} with Xavier initialization. We center crop patches of size 352 × 352 pixels from each image. 
Since Hounsfield unit in CT image is a normalized metric with physical meaning, we do not apply any intensity augmentation.
\item For BraTS20\footnote{BraTS20  contains 16,896 2D slices from MRI scans of 369 patients, with each slice containing a tumor region. For our study, we use only the FLAIR sequence and combine all tumor tissues into a single class.}, we utilize an Inception U-Net \cite{punn2020inception} with Xavier initialization. We use only the FLAIR sequence from the MRI and combine all tumor subclasses into a single "tumor" class. Each slice is natively sized at 240 × 240 pixels, and no cropping or resizing is applied to this dataset.
\end{enumerate}

\subsubsection{Loss Function and Optimization}
The four model-dataset pairs are trained with the following hyperparameters:
\begin{enumerate}
    \item For VOC, the model is trained using a cross-entropy loss (CELoss) with a batch size of 16. The optimizer is Stochastic Gradient Descent (SGD) with a base learning rate of 0.01, momentum of 0.9, and weight decay of 1e-4. A polynomial learning rate scheduling policy is applied, where the learning rate at iteration $t$ is given by: $$lr_{t}=lr_{base}*(1-\dfrac{t}{t_{max}})^{0.9}$$
as formulated by Chen et al. \cite{chen2017deeplab}.
For each (time constraint, annotation style) pair, we validate the model performance after each epoch, saving the weights of the models with the top three validation mIoUs, and report the average of these top three validation performances.
\item For WATER, the model is trained using the Adam optimizer with a learning rate of 1e-3 on CELoss, with no learning rate scheduling. 
A batch size of 16 is used. 
For each (time constraint, annotation style) pair, we validate the model after each epoch, saving the model weights with the highest mean validation IoUs for both background and foreground.
The performance of this saved model is then reported on a held-out test set, which was not used for training or validation.
\item For LiTS17, we use an SGD optimizer with a base learning rate of 0.01, momentum of 0.9, and weight decay of 1e-5. 
The learning rate is adjusted every 3000 iterations by multiplying it by 0.1.
Additionally, we dynamically adjust the weight of each class based on their validation IoU. The relative weight for the cross-entropy loss is computed using the formula: $$\dfrac{1}{\max(0.05,class\ IoU)}$$  
For each (time constraint, annotation style) pair, the model trained at the epoch with the highest mean validation IoU is cached as the final model.
This dataset is used for binary segmentation tasks, and we report the segmentation IoU for the liver class.
\item For BraTS20, the model is trained and cached with the same loss function, batch size, optimizer settings and model selection procedure as WATER.
\end{enumerate}

\subsubsection{Training Framework}
As mentioned in Section \ref{bg}, training models using weak annotations often results in poor performance compared to training with precise annotations. To address this, label pre-processing and annotation-time training are typically required. 
\begin{enumerate}
\item Bounding Box: We examined SDI \cite{khoreva2017simple} and SAM \cite{kirillov2023segment} {(and MedSAM \cite{MedSAM} for medical datasets)} as sub-variants for converting bounding box annotations into segmentation masks, facilitating effective training supervision without modifying the training framework used for precise annotations.  
SDI generates pseudo-masks by combining segmentation results from Multiscale Combinatorial Group (MCG) \cite{arbelaez2014multiscale} and GrabCut \cite{rother2004grabcut} based on the bounding box input. 
Pixels with disagreements between MCG and GrabCut are labeled as the "void" class and excluded from gradient computation. 
{After SDI, various subsequent algorithms have emerged; however, they either demonstrate similar framework or comparable performance to SDI, or were eventually outperformed by SAM in pseudo-mask generation.}
Meanwhile, SAM is based on deep neural networks and can produce instance segmentation results from bounding boxes or point prompts without assumptions about the object's appearance. We use SAM’s inference output as an alternative method for generating pseudo-masks and compare the performance of models trained with these outputs to those trained with segmentation masks generated by SDI.

\item {Scribbles: A common practice in the literature to handle scribble annotations is }by integrating a DenseCRF head \cite{zheng2015conditional}, directly into the model. The DenseCRF head’s loss, known as CRFLoss \cite{lin2016scribblesup}, acts as a regularization term on the model output. This {approach propagates information to unlabeled pixels while simultaneously learning the network parameters.} The final loss function {takes the form}:
$$L_t = \sum_{i\in N}CrossEntropy(Y_t^{(i)},S_t^{(i)})+\lambda CRFLoss(S_t^{(i)})$$
where $\lambda$ is set to $1\times10^{-9}$, as proposed by Tang et al.\cite{tang2018regularized}.
{As an alternative approach, we consider the possibility of combining scribbles with SAM \cite{kirillov2023segment} by sampling pixels taken by the scribble and giving their coordinates as point prompts to SAM. We report the model performance from both variants of model training.}    

\item Points: {Point prompts can be considered as special cases of scribble prompts where each prompt consists of only one pixel. We used the same experiment protocols for both scribbles and points.} In this approach, the segmentation masks with the highest confidence scores from SAM are used as labels for model training.

\end{enumerate}
\section{Results}
To determine the most cost-effective annotation type for segmentation tasks, we vary the annotation time budget and assess model performance across different budget levels. Specifically, the budget is increased by a factor of two until all images in the training set are precisely annotated. If the budget exceeds the number of available images, we train the model using all the labeled data at our disposal.

\subsection{Effects of noise level on model performance}
\label{sanity_check}
Before conducting the main experiment on varying annotation styles within a fixed annotation budget, {We first analyze the impact of annotation quality on model performance. This experiment serves two purposes: first, to determine how errors in noisy masks, presumably created unbiasedly by human annotators, affect the final model performance; and second, to assess the risky of drawing erroneous conclusion due to potentially inaccurate mask quality estimation from  human participants' data.} Here, {we} adjusted the parameters for generating coarse contour labels (rough) to produce variants with mIoUs of [0.81, 0.84, 0.87, 0.90, 0.93] compared to precise annotations. Models were trained on each of these five variants with increasing quality, and the relationship between annotation quality and model performance was quantitatively analyzed using Spearman correlation.
\begin{figure}[ht!]
     \centering
     \begin{subfigure}[b]{0.49\textwidth}
         \raggedright
         \includegraphics[width=\textwidth]{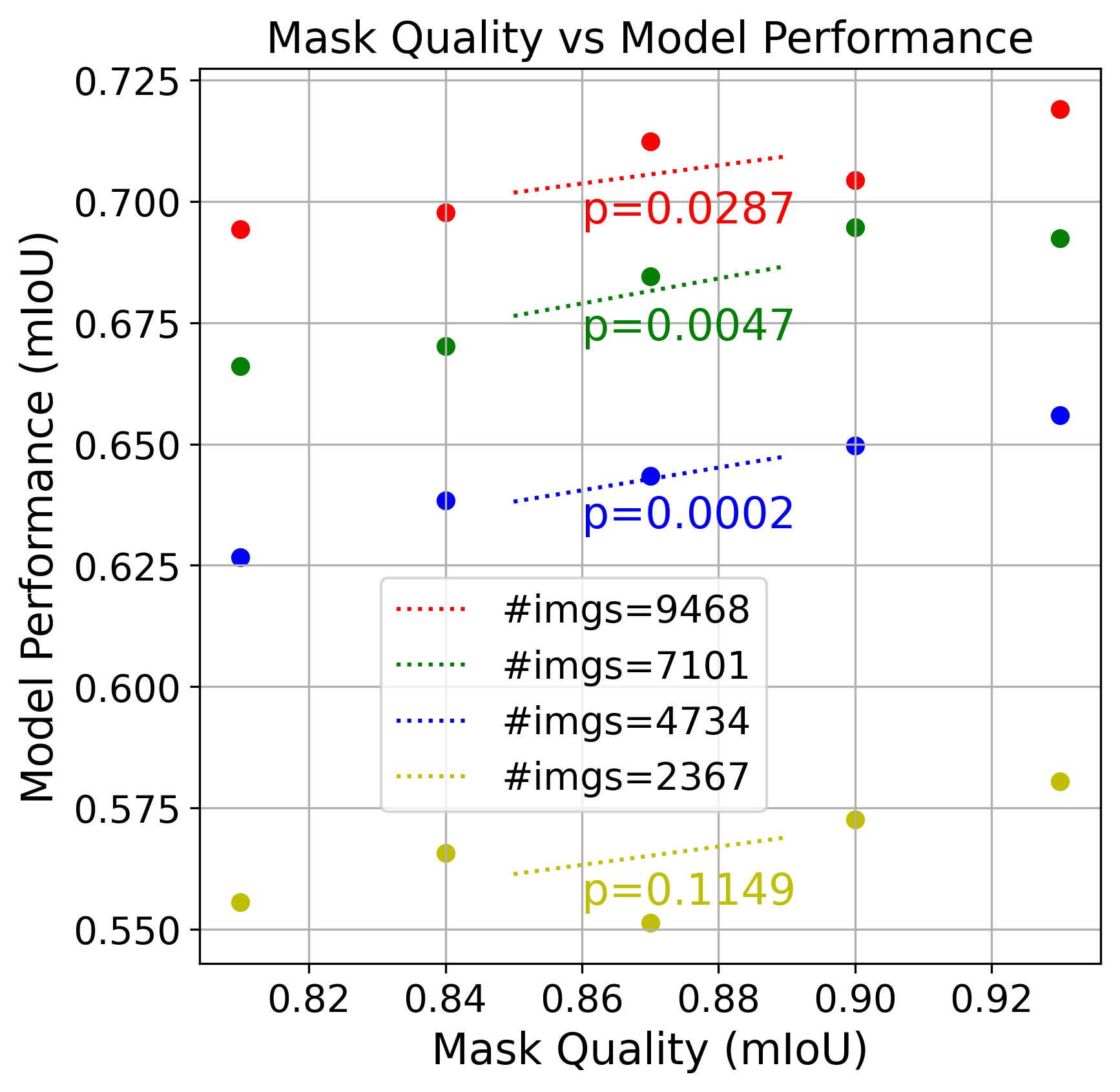}
         \caption{{VOC2012}}
         \label{MQ_VOC}
     \end{subfigure}
     \hfill
     \begin{subfigure}[b]{0.49\textwidth}
         \raggedleft
         \includegraphics[width=\textwidth]{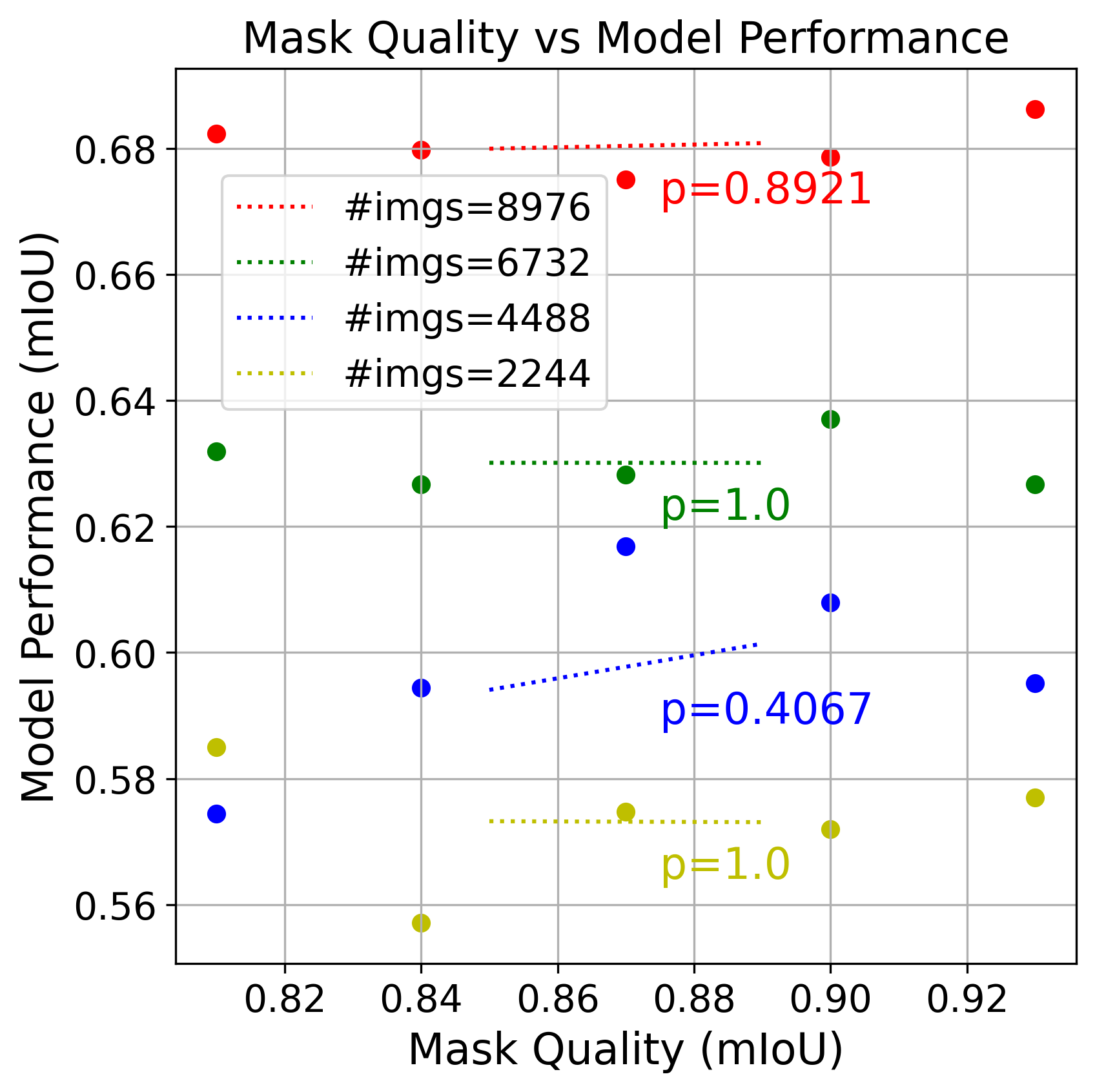}
         \caption{{BraTS20}}
         \label{MQ_Brat}
     \end{subfigure}
     \caption{Effects of Training Mask Quality on Model Performance with Fixed Sample Counts. {To ensure the robustness of this trend, we experiment with the consumption of varying fractions of total costs or budgets, represented as different numbers of annotated images. The dotted line represents general trends for easier visual interpretation and does not correspond to a linear regression. The p-value is calculated using Spearman's rank correlation coefficient.}}
\end{figure}

In VOC2012, we observed a statistically significant (p$\leq0.05$) increase in model performance with higher quality rough labels. However, reducing the mask error rate by half yielded only marginal performance gains compared to doubling the training size. (Fig. \ref{MQ_VOC}) In BraTS20, no statistically significant trend was found across the IoU range (0.81 to 0.93) for any training set size. This lack of significant improvement may be due to the inherent difficulty in precisely defining tumor boundaries from the FLAIR sequence (Fig. \ref{MQ_Brat}). 
{This observation provides us some evidence that the potential inaccuracies in our estimation of mask quality made by human is unlikely to significantly impact the results and lead us to erroneous conclusions on the trade-offs between different annotation styles.}

\subsection{Result on Cost-Effectiveness}
\subsubsection{VOC}
\begin{figure}
    \centering
    \begin{subfigure}[b]{0.49\textwidth}
         \centering
         \includegraphics[width=\textwidth]{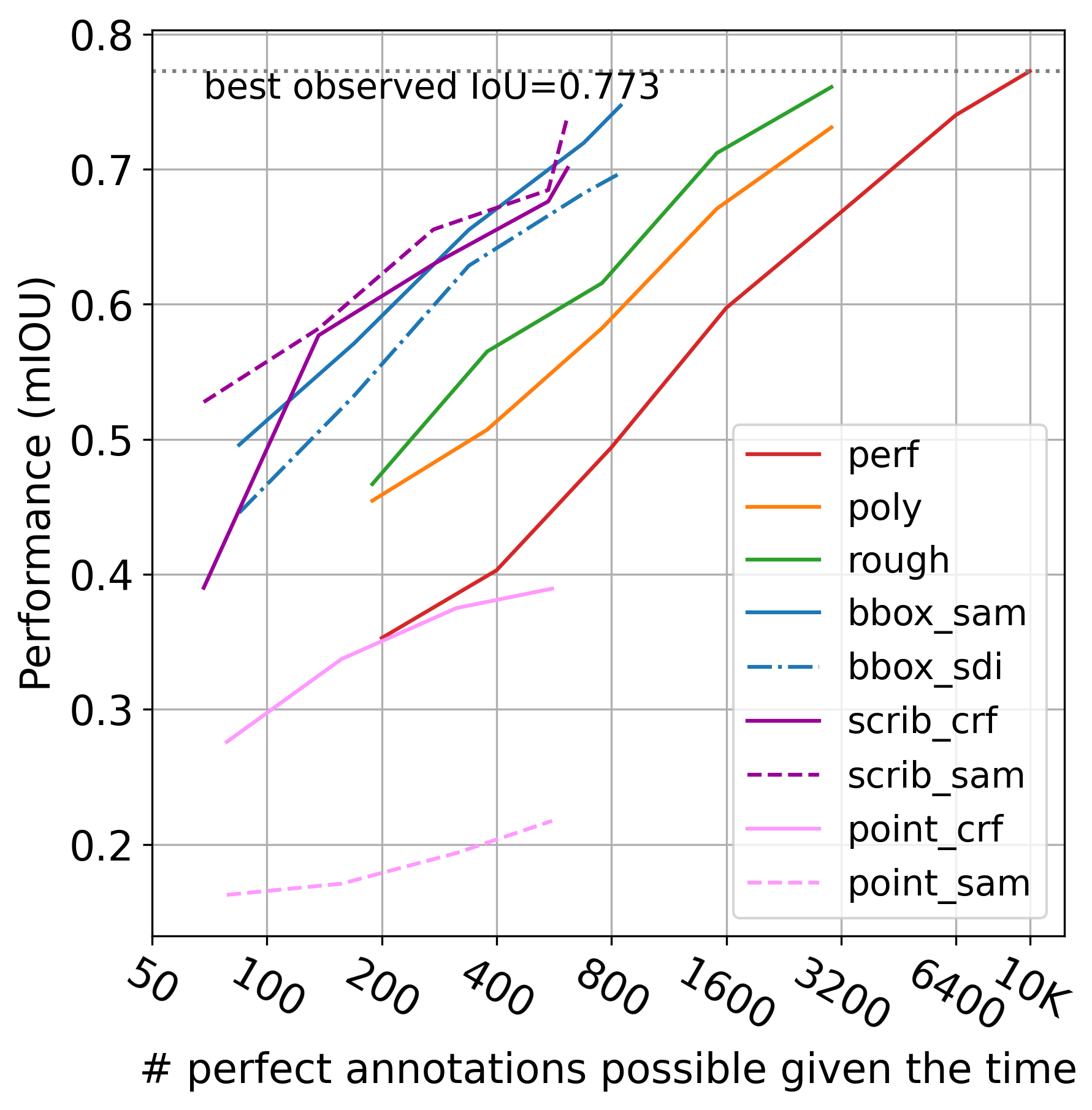}
         \caption{\centering{VOC2012}}
         \label{VOC_full}
    \end{subfigure}
    \hfill
    \begin{subfigure}[b]{0.49\textwidth}
         \centering
         \includegraphics[width=\textwidth]{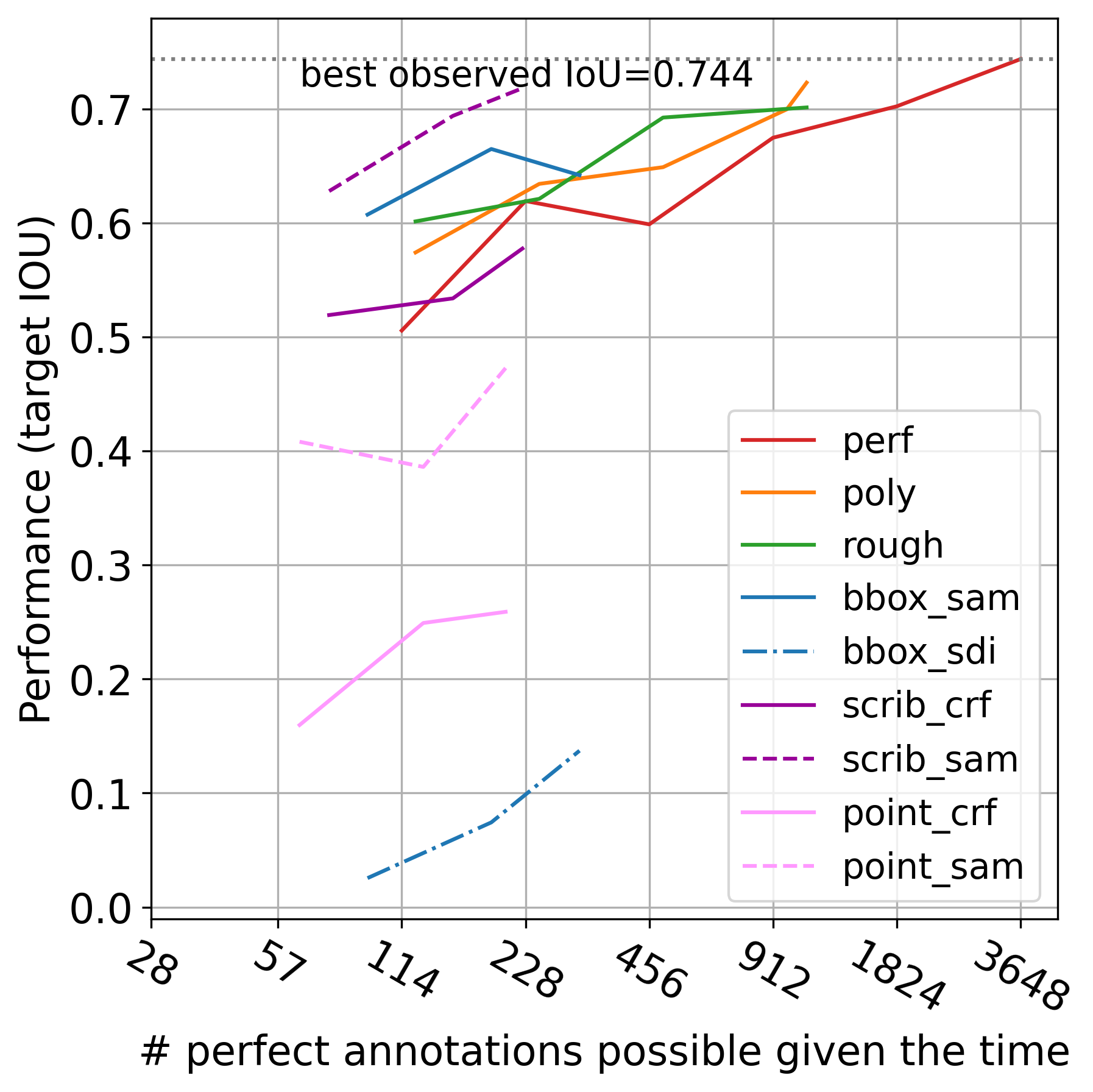}
         \caption{\centering{WATER}}
         \label{WATER_full}
    \end{subfigure} 
    \caption{{The two plots show the approximated cost-effectiveness of different annotation styles for dataset in natural domain. The annotation styles positioned closer to the upper-left corners have higher cost-effectiveness.}}
    
\end{figure}

Experiments on VOC (Fig. \ref{VOC_full}) reveal that most weak or noisy annotation styles are more cost-effective than precise annotations. Among these, SAM-generated masks with bounding box prompts (bbox\_sam) exhibit the highest cost-effectiveness, followed by coarse contour annotations (rough), reflecting SAM’s strong performance when paired with bounding box prompts. However, SAM with point prompts {yielded a surprisingly low performance}. We found that SAM's highest-confidence labels from point prompts often cover only a small part of the target region. For instance, if a point is placed on a car window, only that specific part of the window is marked, missing the broader target area. This limitation stems from the ambiguity of a single point and the per-contiguous-region nature of point prompts, which reduces their effectiveness in distinguishing multiple instances. {In contrast, the multi-point prompts derived from scribbles + SAM counterpart achieved similarly good cost-effectiveness to bbox\_sam.} Although point annotations with iterative training demonstrate slightly higher cost-effectiveness than point\_sam, they still rank lowest among the annotation styles tested.

\subsubsection{WATER}
Like in VOC, {polygon (poly)} and coarse contour (rough) annotations consistently exhibit better cost-effectiveness than precise annotations. Bounding box masks generated with SDI failed to produce a model with satisfactory predictive capability. This may be due to the {nature of the target in WATER:} unlike VOC, where ROIs are typically {having} complete and often convex {shape}, the WATER dataset focuses on “stuff” {that contain highly irregularly-shaped ROIs, which may be beyond the original usage case of SDI.} In contrast, SAM showed decent initial performance with bounding box prompts, though it quickly plateaued. {Point prompt plus SAM may be similarly affected. For scribbles, without leveraging foundation models, integrating the CRF Loss as a regularization term still achieved moderate cost-effectiveness. Interestingly, SAM combined with scribbles outperformed all other methods, achieving both the highest cost-effectiveness and the best full-dataset performance for this dataset. A single point may be ambiguous for "stuff", but a list of points sampled from scribbles are well-suited for it.}
 (See Fig. \ref{WATER_full})

\subsubsection{LiTS17}
\begin{figure}
    \centering
    \begin{subfigure}[b]{0.49\textwidth}
         \centering
         \includegraphics[width=\textwidth]{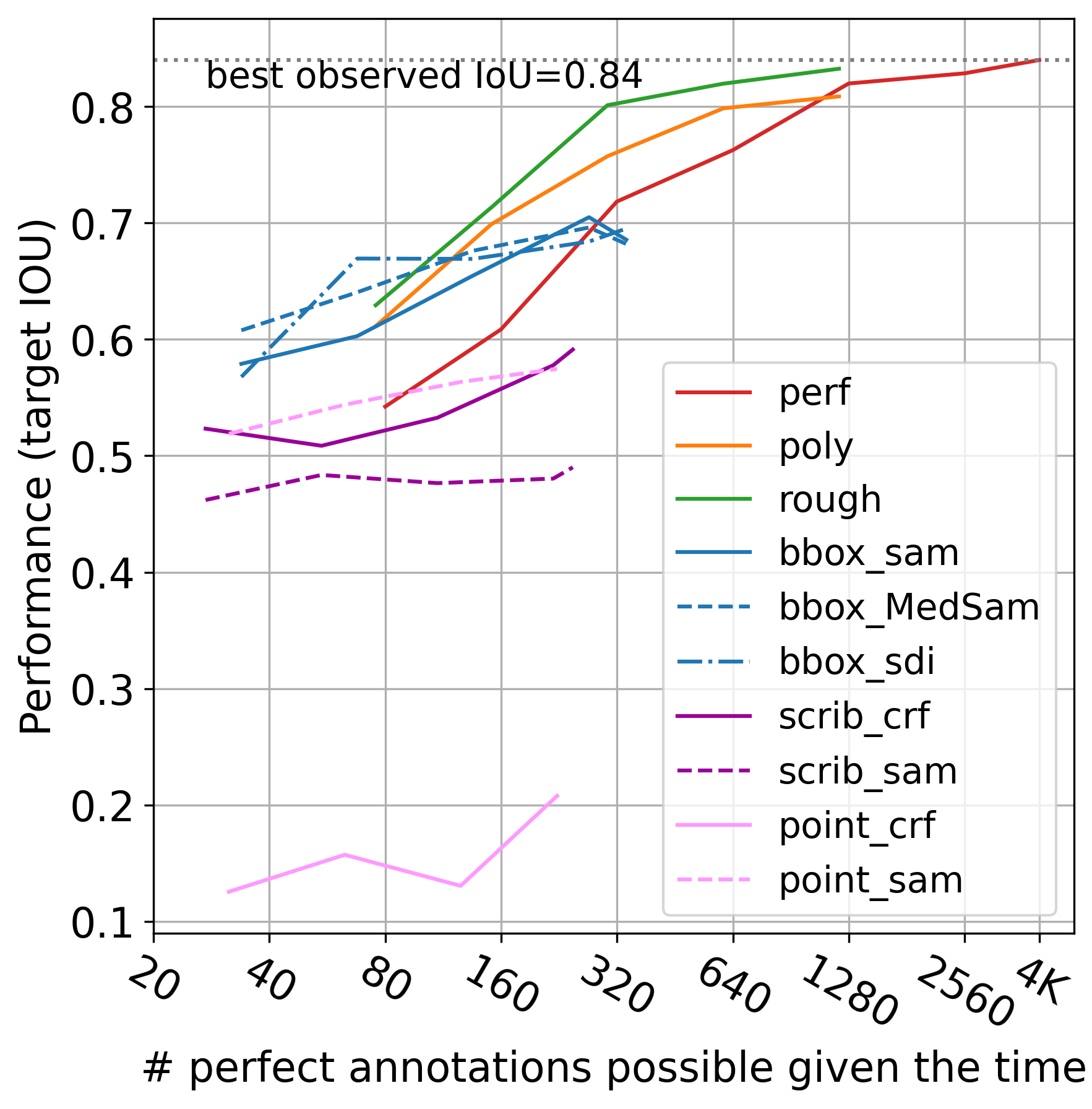}
         \caption{{LiTS17 (CT)}}
         \label{LiTS_full}
    \end{subfigure}
    \hfill
    \begin{subfigure}[b]{0.49\textwidth}
         \centering
         \includegraphics[width=\textwidth]{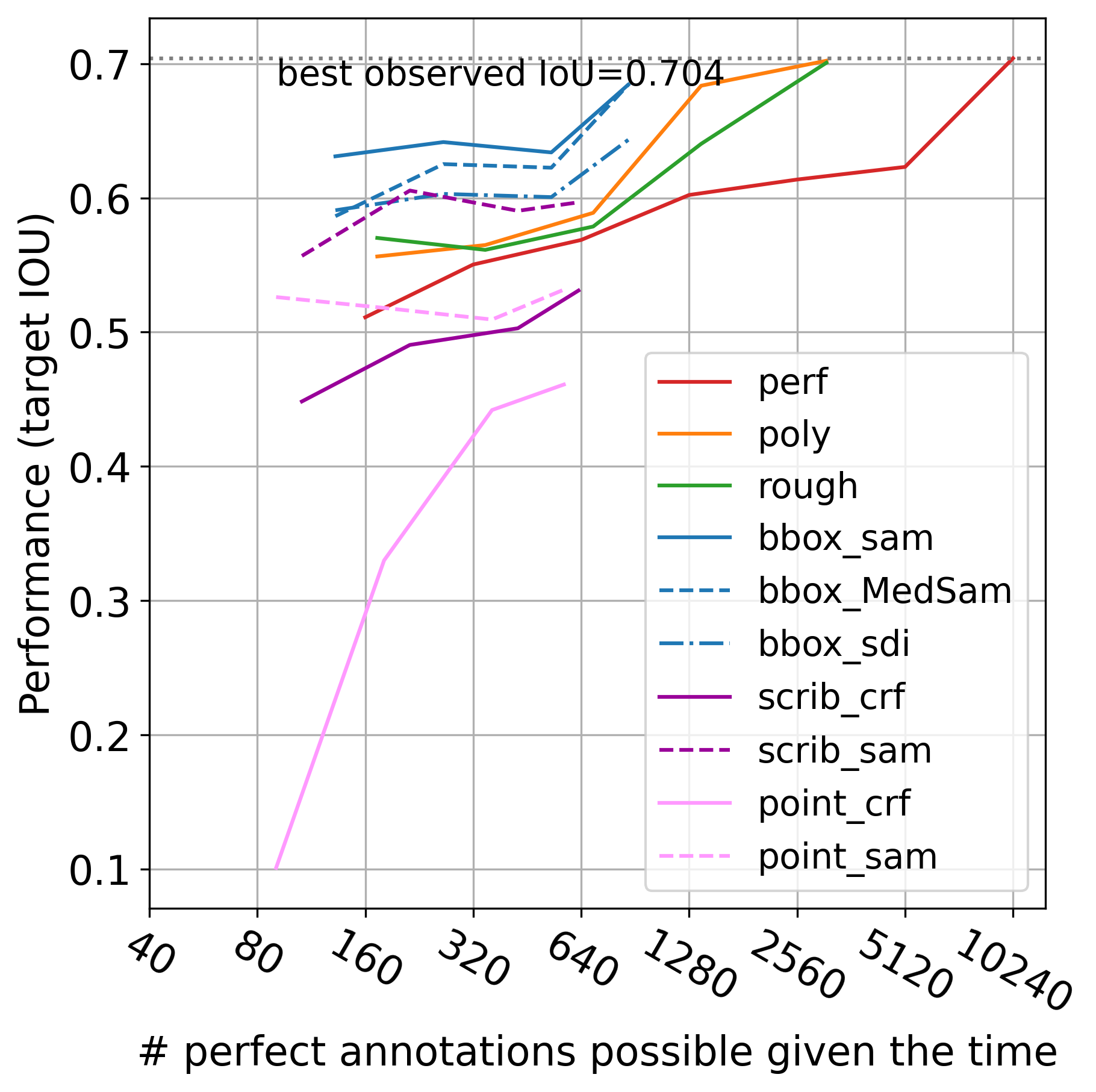}
         \caption{{BraTS20 (MRI)}}
         \label{BraTS_full}
    \end{subfigure} 
    \caption{{The two plots show the approximated cost-effectiveness of different annotation styles for dataset in CT and MRI. both SAM and MedSAM \cite{MedSAM} are included as candidate model for processing bounding box annotations.}}
    
\end{figure}

For LiTS17,  {polygon (poly) and coarse contour (rough)} labels consistently show higher cost-effectiveness than precise labels. Coarse contour labels, in particular, lead to better model performance compared to both high-precision and coarse polygon annotations as available annotation time increases. This improved performance may be due to the smooth boundaries of coarse contours, which more closely match the liver boundaries than vertex-based annotations. Models trained {with the three variants of bounding box annotations generated similar cost-effectiveness curves, yet are no longer more cost-effective than the noisy annotations (presumably delineated by humans annotators).}   Scribble and point annotations demonstrate even lower cost-effectiveness. {Anti-intuitively, we observed scrib\_sam to have lower performing than point\_sam.} (See Fig. \ref{LiTS_full})

\subsubsection{BraTS20}
For BraTS20, models trained with {both polygon and rough contour annotations achieve comparable}  maximum performance to those trained with precise annotations. Both {SAM and MedSAM-augmented} bounding box annotations demonstrate {similarly} high cost-effectiveness, {followed by the SDI-augmented counterpart. Similar to what's observed in LiTS17,} point and scribble annotations {with crf} exhibit lower cost-effectiveness and performance {than all other annotation styles (Fig. \ref{BraTS_full})}. {Scrib\_sam yields a higher cost-effectiveness than the noisy annotations, yet a lower performance when all images are included.}

Through experiments assessing cost-effectiveness across these four datasets, we observe the following common phenomena:
\begin{enumerate}
\item {Pixel-wise accurate annotation is rarely the most cost-effective annotation style for semantic segmentation, given a fixed annotation budget.}
\item Using noisy annotations with reasonable annotation quality does not significantly compromise model performance when compared to precise annotations for the same set of training images.
\item For labeling object-like ROIs, employing bounding box annotations followed by post-processing with zero-shot segmentation networks (e.g., SAM, RITM, SimpleClick) can be a highly cost-effective data-labeling strategy.
\item {Noisy annotations (i.e., polygon and coarse contour) remain to be relatively cost-effective annotations style for some datasets or segmentation tasks, despite the availability of prompt-based foundation models.}
\item {Scribbles (or multi-point annotations) appear to be highly cost-effective annotation style for natural-domain images, yet their performance when all images are consumed is sub-optimal, if not undesirable, for medical-domain images.}

\end{enumerate}
\section{Discussion}
In our study, we evaluated the cost-effectiveness of various weak and noisy annotation styles for training segmentation networks. We found that, within a reasonable range of annotation quality (mIoU $\geq$ 0.8), the effect of annotation quality on model performance is minimal. Across all four datasets, {there exists multiple variants of weak and noisy annotation which obtain higher cost-effectiveness and comparable full-dataset performance to the precisely annotated counterpart.} We also found that single-point annotations combined with SAM {performed no better than scribble annotations with CRF Loss both in terms of  cost-effectiveness and full-dataset performance for natural images. This may be due to the ambiguity in the target reference associated with single-point annotations. For medical images, however, the two methods performed similarly.}{On the other hand, multi-point prompts sampled from scribbles emerged as a highly cost-effective and high-performing annotation style for natural-domain images with comparable or even higher performance than bbox\_sam. However, for medical-domain images, this observation flips. More investigations will be needed to full understand the cause of this disparity.} 

Our study has some limitations: First,
we assumed a single, homogeneous annotation style for model training {and did not incorporate Semi-Supervised Learning (SSL) technique to leverage unannotated images.} In practice, {researchers can select} mixed annotation styles and {utilize a training framework supporting the use of a mixed type of annotations styles. However, evaluating existing mixed supervision models and determining the optimal ratio for each annotation type are beyond the scope of this study. This uncovered topic is a promising direction for future research.}
{Second, we reported results using a train-validation-test experimental setup rather than an N-fold cross-validation approach. The specific samples selected may influence the numerical outcomes of model performance, potentially leading to quantitative inaccuracies. However the substantial differences observed across annotation styles suggest that the qualitative conclusions drawn in this study remain robust.}

{
Our study points to several potential topics for future research: (1) open-source foundation models capable of accepting a mixed type of imperfect annotation styles as input, (2) implications of using imperfect (or sparse) annotations on 3D medical modalities, such as volumetric MRI and CT scans, and (3) SAM extensions providing broader supports to point and scribble prompts, especially for images in medical-domains.}

\section{Conclusion}
{Our study evaluates the cost-effectiveness of weak and noisy annotation styles for training segmentation networks. We found that pixel-wise accurate annotations are rarely the most cost-effective option. However, noisy annotations such as polygons and coarse contours, and bounding boxes, when paired with models like SAM, offer cost-effective alternatives under many circumstances. Scribbles are highly effective for natural images but perform sub-optimally in medical imaging -- possibly due to domain-specific challenges. Models similar to SAM but with better supports for scribble prompts and medical images might help mitigate this issue. We recommend researchers to consider the option of using imperfect annotation, when appropriate, to make the best use of the annotation budget in their research funding.}


\bibliography{sn-bibliography}
\clearpage

\appendix
\newpage%
\section{Appendix: Quotes from Annotation Providers}
\begin{table}[h!]
\centering
\begin{tabular}{|l|l|l|l|l|l|}
\hline
\textbf{Vendor} & \textbf{precise} & \textbf{polygon} & \textbf{bounding box} & \textbf{point} & \textbf{scribble} \\
\hline
Vendor A & 0.3 & 0.06 & 0.04 & 0.03 & 0.03 \\
Vendor B & 850 & 180 & 49 & 25 & - \\
Vendor C & 12 & 8 & 5 & - & - \\
Vendor D & 50 & 35 & 12 & - & - \\
Vendor E & 40 & 7 & 3 & - & - \\
Vendor F & 870 & 257 & 63 & - & - \\
Vendor G & 0.5 & 0.08 & - & 0.04 & 0.04 \\
Vendor H & 6 & 3 & - & 1 & - \\
Vendor I & 0.84 & - & 0.036 & 0.012 & 0.036 \\
Vendor J & 0.84 & - & 0.035 & 0.012 & 0.036 \\
\hline
\end{tabular}
\caption{Original Value from Vendors' Quotes}
\label{vendors_original}
\end{table}

\begin{table}[h!]
\centering
\begin{tabular}{|l|l|l|l|l|l|}
\hline
\textbf{Vendor} & \textbf{perfect} & \textbf{polygon} & \textbf{bounding box} & \textbf{point} & \textbf{scribble} \\
\hline
Vendor A & 1 & 0.2 & 0.1333 & 0.1 & 0.1 \\
Vendor B & 1 & 0.2118 & 0.0576 & 0.0294 & - \\
Vendor C & 1 & 0.6667 & 0.4167 & - & - \\
Vendor D & 1 & 0.7 & 0.24 & - & - \\
Vendor E & 1 & 0.175 & 0.075 & - & - \\
Vendor F & 1 & 0.2954 & 0.0724 & - & - \\
Vendor G & 1 & 0.16 & - & 0.08 & 0.08 \\
Vendor H & 1 & 0.5 & - & 0.1667 & - \\
Vendor I & 1 & - & 0.0429 & 0.0143 & 0.0429 \\
Vendor J & 1 & - & 0.0417 & 0.0143 & 0.0429 \\
\hline
\end{tabular}
\caption{Cost Ratio for Different Annotation Styles}
\label{vendors_normalized}
\end{table}
The original quotes from the 10 annotation vendors are reported in Table. \ref{vendors_original}. and the cost ratios among different annotation styles are computed in Table. \ref{vendors_normalized}. After taking a $50\%$ trimmed mean for the ratio for each imperfect annotation style, the relative cost ratio is reported in Table. \ref{mask_cost_and_quality}.

\newpage
\section{Appendix: Examples of Computer-Generated Imperfect Annotations}
\begin{figure}[ht!]
    \centering
    \begin{subfigure}[b]{0.32\textwidth}
        \centering
        \includegraphics[width=\textwidth]{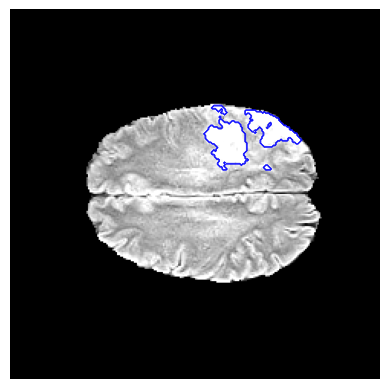} 
        \caption{Precise (prec)}
    \end{subfigure}
    \begin{subfigure}[b]{0.32\textwidth}
        \centering
        \includegraphics[width=\textwidth]{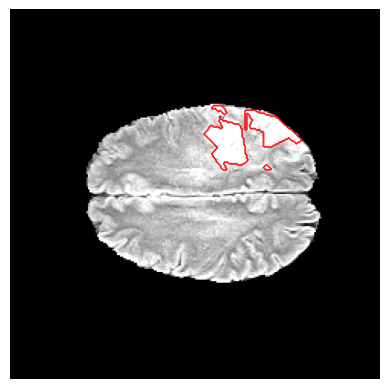} 
        \caption{Polygon (poly)}
    \end{subfigure}
    \begin{subfigure}[b]{0.32\textwidth}
        \centering
        \includegraphics[width=\textwidth]{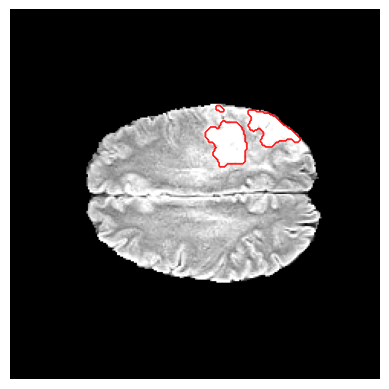} 
        \caption{Coarse Contour (rough)}
    \end{subfigure}
    \begin{subfigure}[b]{0.32\textwidth}
        \centering
        \includegraphics[width=\textwidth]{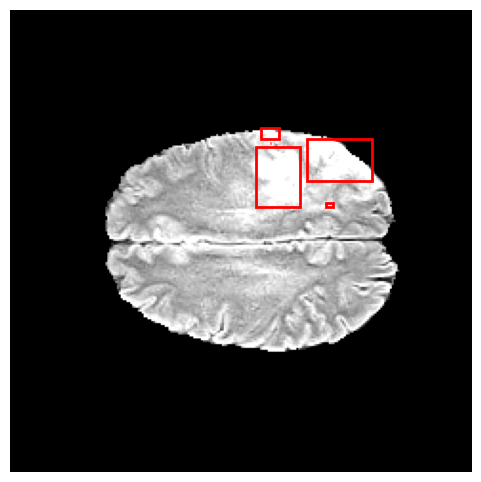} 
        \caption{Bounding Box (bbox)}
    \end{subfigure}
    \begin{subfigure}[b]{0.32\textwidth}
        \centering
        \includegraphics[width=\textwidth]{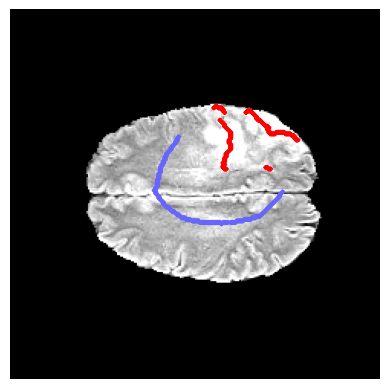} 
        \caption{Scribble (scrib)}
    \end{subfigure}
    \begin{subfigure}[b]{0.32\textwidth}
        \centering
        \includegraphics[width=\textwidth]{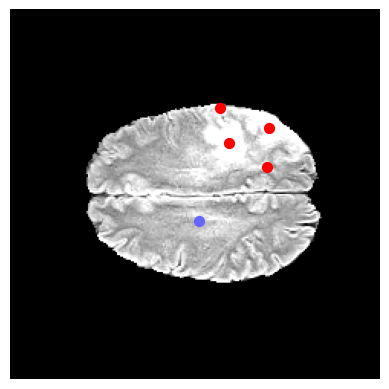} 
        \caption{Point (point)}
    \end{subfigure}
    \caption{{Computer-generated annotations that assimilate the six different
annotation styles on BraTS20. }}
    \label{BraTS20_annot_samples}
\end{figure}

{The visualization of datasets other than VOC2012 is listed  below in Fig. \ref{WATER_annot_samples},\ref{LiTS17_annot_samples},\ref{BraTS20_annot_samples}}
\begin{figure}[ht!]
    \centering
    \begin{subfigure}[b]{0.32\textwidth}
        \centering
        \includegraphics[width=\textwidth]{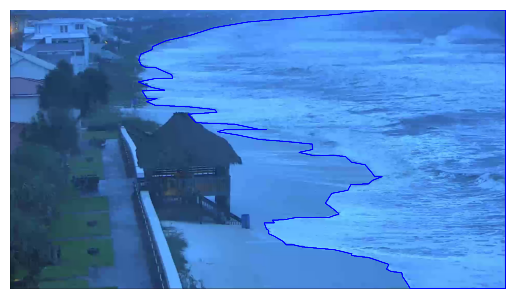} 
        \caption{Precise (prec)}
    \end{subfigure}
    \begin{subfigure}[b]{0.32\textwidth}
        \centering
        \includegraphics[width=\textwidth]{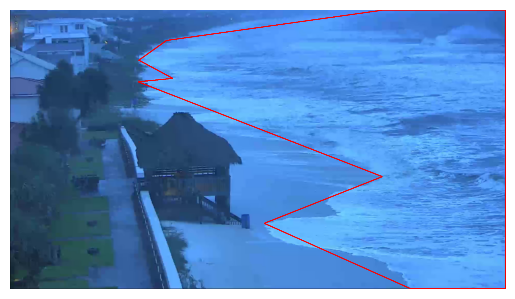} 
        \caption{Polygon (poly)}
    \end{subfigure}
    \begin{subfigure}[b]{0.32\textwidth}
        \centering
        \includegraphics[width=\textwidth]{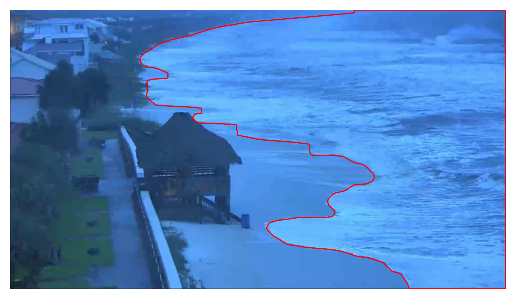} 
        \caption{Coarse Contour (rough)}
    \end{subfigure}
    \begin{subfigure}[b]{0.32\textwidth}
        \centering
        \includegraphics[width=\textwidth]{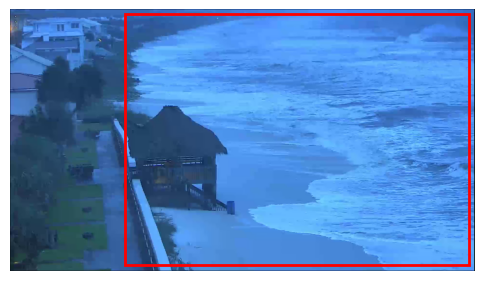} 
        \caption{Bounding Box (bbox)}
    \end{subfigure}
    \begin{subfigure}[b]{0.32\textwidth}
        \centering
        \includegraphics[width=\textwidth]{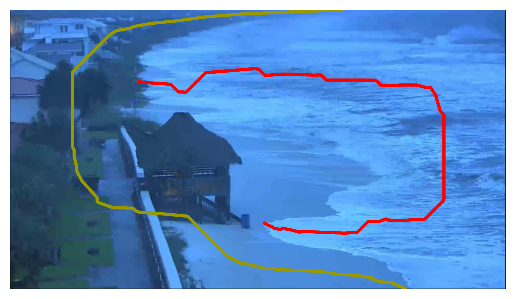} 
        \caption{Scribble (scrib)}
    \end{subfigure}
    \begin{subfigure}[b]{0.32\textwidth}
        \centering
        \includegraphics[width=\textwidth]{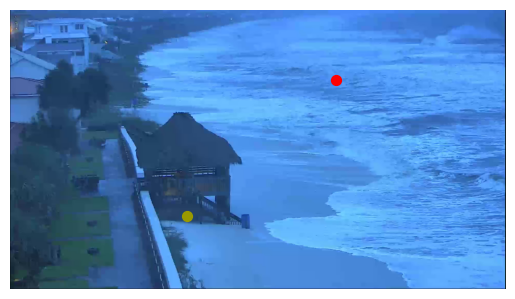} 
        \caption{Point (point)}
    \end{subfigure}
    \caption{{Computer-generated annotations that assimilate the six different
annotation styles on WATER.}}
    \label{WATER_annot_samples}
\end{figure}

\begin{figure}[h]
    \centering
    \begin{subfigure}[b]{0.32\textwidth}
        \centering
        \includegraphics[width=\textwidth]{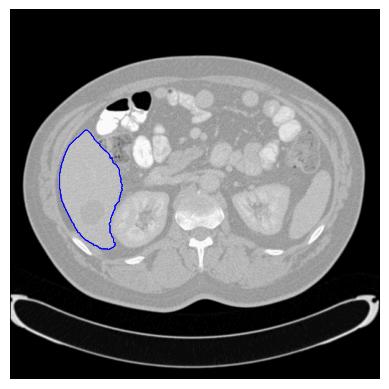} 
        \caption{Precise (prec)}
    \end{subfigure}
    \begin{subfigure}[b]{0.32\textwidth}
        \centering
        \includegraphics[width=\textwidth]{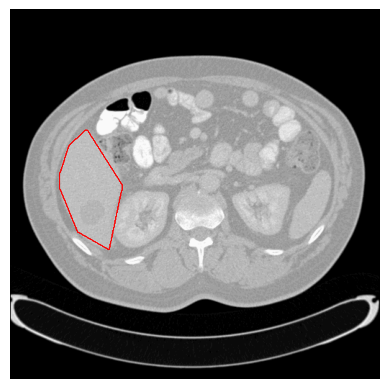} 
        \caption{Polygon (poly)}
    \end{subfigure}
    \begin{subfigure}[b]{0.32\textwidth}
        \centering
        \includegraphics[width=\textwidth]{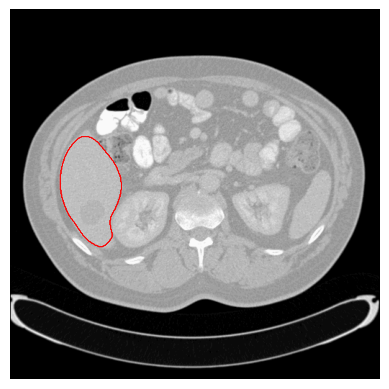} 
        \caption{Coarse Contour (rough)}
    \end{subfigure}
    \vspace{-1mm}
    \begin{subfigure}[b]{0.32\textwidth}
        \centering
        \includegraphics[width=\textwidth]{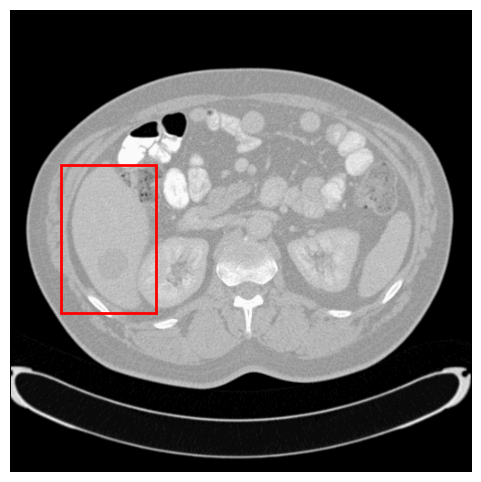} 
        \caption{Bounding Box (bbox)}
    \end{subfigure}
    \begin{subfigure}[b]{0.32\textwidth}
        \centering
        \includegraphics[width=\textwidth]{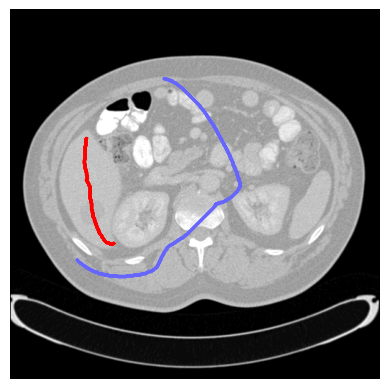} 
        \caption{Scribble (scrib)}
    \end{subfigure}
    \begin{subfigure}[b]{0.32\textwidth}
        \centering
        \includegraphics[width=\textwidth]{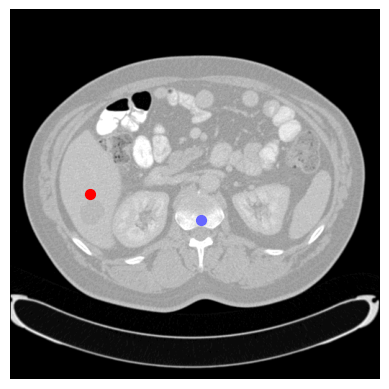} 
        \caption{Point (point)}
    \end{subfigure}
    \caption{{Computer-generated annotations that assimilate the six different
annotation styles on LiTS17. }}
    \label{LiTS17_annot_samples}
\end{figure}

\end{document}